\documentclass{article}

\usepackage{arxiv}

\usepackage[utf8]{inputenc} 
\usepackage[T1]{fontenc}    
\usepackage{hyperref}       
\usepackage{url}            
\usepackage{booktabs}       
\usepackage{amsfonts}       
\usepackage{nicefrac}       
\usepackage{microtype}      
\usepackage{lipsum}		
\usepackage{graphicx}
\usepackage{natbib}
\usepackage{doi}
\usepackage{wrapfig}
\usepackage{svg}
\usepackage{amsmath}

\title{Cognitive Modeling of Semantic Fluency Using Transformers}

\date{} 					

\author{Animesh Nighojkar, Anna Khlyzova, John Licato \\
	Advancing Machine and Human Reasoning (AMHR) Lab \\
	Department of Computer Science and Engineering \\
	University of South Florida, Tampa, USA \\
	\texttt{(anighojkar, annakhlyzova, licato)@usf.edu}
}

\renewcommand{\undertitle}{}
\renewcommand{\headeright}{}

\hypersetup{
pdftitle={Cognitive Modeling of Semantic Fluency Using Transformers},
pdfsubject={q-bio.NC, q-bio.QM},
pdfauthor={Animesh Nighojkar, Anna Khlyzova, John Licato},
}

\begin{document}
\maketitle

\begin{abstract}
	Can deep language models be explanatory models of human cognition? If so, what are their limits? In order to explore this question, we propose an approach called \textit{hyperparameter hypothesization} that uses predictive hyperparameter tuning in order to find individuating descriptors of cognitive-behavioral profiles. We take the first step in this approach by predicting human performance in the semantic fluency task (SFT), a well-studied task in cognitive science that has never before been modeled using transformer-based language models (TLMs). In our task setup, we compare several approaches to predicting which word an individual performing SFT will utter next. We report preliminary evidence suggesting that, despite obvious implementational differences in how people and TLMs learn and use language, TLMs can be used to identify individual differences in human fluency task behaviors better than existing computational models, and may offer insights into human memory retrieval strategies---cognitive process not typically considered to be the kinds of things TLMs can model. Finally, we discuss the implications of this work for cognitive modeling of knowledge representations.
\end{abstract}


\section{Introduction}
\label{sec:intro}

Two of the most important ideas underpinning contemporary cognitive science--and the closely related AI subfield of computational cognitive modeling--are the suppositions that the human mind uses cognitive structures and that progress in understanding the mind can come from modeling those structures and the algorithms which operate on them.
The semantic fluency task (SFT), sometimes called the verbal fluency task \cite{welsh1991normative},
is commonly employed in service of those goals. In SFT, participants name as many items belonging to a particular semantic category (animals, fruits, etc.) as they can in a fixed amount of time (typically 40-180 seconds).
Despite this task's simplicity, the lists generated by participants (which we call semantic fluency lists or SFLs) offer insights into the structure of human knowledge and the heuristics used for memory retrieval. For example, words sharing semantic features tend to group in clusters, and there is often a temporal delay before a participant switches from one cluster to another.

Multiple approaches to computationally modeling behaviors in SFT have been proposed
\cite{hills2012optimal,abbott2015random,zemla2016u,zemla2017modeling,avery2018comparing}, 
most relying on graph-based representations in which words are represented as nodes, and edges correspond to some meaningful semantic relationship between the nodes.
However, \textit{to date, no work has explored whether transformer-based language models (TLMs) can be any better at modeling the generation of SFLs}. And there are multiple 
reasons, at least from an exploratory perspective, to suspect TLMs might do well in this regard, e.g.: (1) a large body of literature demonstrates why semantic memory can not be sufficiently represented purely by fixed associative links between lexical nodes---at minimum, representations must allow for dynamic role binding, hierarchical (or otherwise unidirectional) activations, and enough richness to carry out structure-sensitive similarity assessments 
\cite{Holyoak2000,Sun2002}; (2) TLMs perform unexpectedly well on human-oriented linguistic benchmarks \cite{Wang2019b}, and they are typically pre-trained using a lengthy process designed to embed deep semantic knowledge, resulting in a dense encoding of semantic relationships 
\cite{Cui2020}; (3) The pre-training process often proceeds by optimizing LMs to perform well on the \textit{MLM} (masked language modeling) task, which shares more than a passing resemblance to the kind of word prediction that some researchers believe children are performing 
\cite{Gambi2020}; and (4) TLMs tend to outperform other approaches in recent work modeling human reading times, eye-tracking data, and others 
\cite{Schrimpf2020}. 

Considered altogether, these reasons are sufficient to motivate an initial exploration into TLM-based semantic fluency modeling. Our \textbf{novel contributions} include:
\begin{itemize}
    \item We are the first, to our knowledge, to generate and model SFLs using TLMs; we use RoBERTa-Large, DistilBERT, and miniBERTa-med-small in this paper to further the state-of-the-art on modeling SFLs. Generally, our models significantly outperform more traditional, semantic network-based approaches.
    
    
    
    \item 
    We design two adaptive approaches that predict the next SFL item as they learn from the previous items and turn out to be superior to other non-adaptive approaches. These adaptive approaches, we believe, can serve as baselines against which to compare future computational cognitive models.

    \item In a broader sense, we argue and demonstrate that TLMs, despite being pre-trained using techniques and datasets very different than those human beings use, can be powerful tools for studying human cognition, knowledge representation, and memory retrieval. This is a first step in a computational cognitive modeling strategy that we call \textit{hyperparameter hypothesization} (\S \ref{sec:CMs}).
    
\end{itemize}

Any performance on modeling SFLs discussed in this paper is for a pre-trained model with no fine-tuning on the SFT. We do this because the objective of this work is not to learn how to perform the SFT in the most precise way; it is to model human SFLs in an attempt to use the best performing hyperparameters to learn something about human cognitive traits.


\subsection{Deep Learning as Cognitive Model}
\label{sec:CMs}

Cognitive Science has long benefited from computational cognitive models (CMs), which are computational implementations of cognitive processes, at various levels of abstraction, created typically in order to test theoretical claims \cite{Sun2008}. Furthermore, because carrying out empirical studies with people can involve difficult logistics, the existence of myriad confounding variables, and prohibitive costs, the use of well-designed CMs can save psychologists an immense amount of time and resources, e.g.\ by making it easier to test hypotheses about cognitive processes with CMs prior to empirical work. However, there are fundamental hardware and implementational differences between human brains and silicon-based electronics, raising the question: to what extent can a CM support or refute a theory of human cognition? Although there is a longstanding debate about to which degree the algorithms used by a CM commits it to certain claims about the cognitive phenomena it purports to model (e.g., see \cite{Jones2011,Marcus2012,Marcus2015}), most agree that the level of abstraction the CM represents does entail some \textit{ontological commitment} \cite{Bricker2016,Floridi2011},\footnote{A model that is \textit{ontologically committed} tells us something about the real-world object or process that it is meant to model, rather than simply to match the data that the object or process outputs.} much like any other scientific theory can be said to model and explain something about the natural world. 
In other words, if we want a CM to be able to teach us something about the human mind, its design choices cannot be made arbitrarily because the way the model works must have some correspondence to the cognitive process it purports to model. How then can massive transformer-based language models, that are trained on large datasets using algorithms and data structures that appear fundamentally different from those used by people, tell us something about human minds? 

Our answer to this important question is brief: 
We propose a technique we call \textit{hyperparameter hypothesization}, the form of which goes as follows: If, for certain values of hyperparameters $\mathcal{H}$: (1) a CM matches large amounts of human data significantly better than other models; (2) the human data matched ranges across a variety of tasks given values of $\mathcal{H}$; and (3) all $h \in \mathcal{H}$ have functional roles in the CM that reasonably align with functional roles known to exist in human cognition; then we can reasonably use it to form a hypothesis about human cognition. For example, suppose we have a TLM with a hyperparameter $w$, which restricts the amount of information that the CM can consider simultaneously. We then find that certain values of $w$ allow the CM to match human data on a cognitive task much better than existing models (e.g., SFT). We may then find that a similar range of values for $w$ allows the CM to match human data on other cognitive tasks as well. This can allow us to reasonably \textit{hypothesize} (but not yet definitively conclude) that this range of values for $w$ corresponds to a similar range in people---we might predict that it corresponds to the amount of \textit{working memory} that people typically have. This hypothesis can then be tested: we can observe how our CM performs on values of $w$ lower than the optimal range and see whether its resulting behaviors align with those of humans who are known to have lower working memory sizes.

Although the above is only one example of how hyperparameter hypothesization may work, its first step involves demonstrating that a certain type of CM can indeed match human data significantly better than others. The remainder of this paper restricts its focus to that, specifically on the semantic fluency task.

\section{Related Work}
\label{sec:related}

Prior attempts to model semantic fluency have largely been based on \textit{semantic networks}, or graph representations where words are nodes and relationships between those nodes are edges. 
At least since \citet{collins1969retrieval}, semantic networks have been a common tool in computational modeling \cite{hills2015foraging,avery2018comparing}, typically using graph representations drawing from
large-scale databases such as WordNet \cite{miller1998wordnet}, text corpora \cite{macwhinney2000childes}, and the USF free association norms \cite{Nelson2004}. The U-INVITE model \cite{zemla2016u} reconstructs individuals' semantic networks using a combination of large-scale databases and semantic fluency data.

Information obtained from analyzing semantic fluency lists (SFLs) can be used to construct portions of semantic networks. But since the amount of data in SFLs is very small, it is more common to instead obtain word association data from larger semantic datasets. For instance, the USF Free Association Norms \cite{Nelson2004} is a free association dataset collected from more than 6,000 participants who were asked to write the first word $w$ that came to their mind given a ``cue word'' $C$. This dataset offers more than 72,000 word pairs ($C$, $w$) along with the percentage of participants who wrote $w$ given $C$.
\citet{zemla2017modeling} used the USF norms to construct a semantic network and simulate a variety of memory search processes, including censored random walk, whose simulations are compared to the results of the previously collected human data in an SFT \cite{zemla2016u}. Small World of Words (SWOW) \cite{dedeyne_navarro_perfors_brysbaert_storms_2019} is a more recent word association dataset offering more than 1.3M ($C$, $w$) pairs. SNAFU \cite{zemla2020snafu} is a tool for estimating semantic networks and analyzing fluency data (including random walk); the authors provide a sample dataset of an SFT called ``SNAFU Sample''.{\footnote{\url{https://github.com/AusterweilLab/snafu-py/blob/master/fluency_data/snafu_sample.csv}}} gathered from 82 participants that contains 796 lists spanning across 6 categories.\footnote{The number of items in each category are \textit{fruits (60), vegetables (60), animals (296), supermarket items (81), tools (149), foods (150)}. The median list lengths are \textit{fruits (18), vegetables (17.5), animals (34), supermarket items (35), tools (16), foods (36.5)}} In this paper, we try to model the SFLs from this dataset.

\citet{hills2012optimal} compare the memory search process to the strategies animals use when searching for food (optimal foraging).
This includes a dynamic process of switching from local search of a cluster of semantically similar items, to a global search when the difficulty of finding an item nearby reaches a certain point. This process is called ``patch switching''. To replicate the dynamic process of switching between patches, the authors implemented a dynamic model that used the previous item recalled and frequency to perform the switching. The model produced a log-likelihood fit, which was then compared to the static models that ignored the patchy structure of the network. The dynamic model showed better results, suggesting that humans perform memory search using patch switching too.

\citet{kajic2017biologically} proposed a biologically-constrained spiking neural network model to produce human-like SFLs. Three different sources of associative data, including the USF norms, were used to construct association matrices for a neural network. To compare the results with the human data, the authors recorded word responses as decoded vector representations and inter-item response times between the adjacent retrieved words. The locality shown in \cite{hills2012optimal} is supported by the results of these experiments: the preceding word is the most similar to the current word in a patch.

A related task is Entity Set Expansion (ESE) \cite{zhang2017} that takes a set of entities as input (and not a category word like SFT) and tries to add new entities to that set after predicting a category all those entities belong to (this additional step is absent in SFT). The fundamental difference between ESE and the work presented in this paper is that we are trying to \textit{model human SFLs} instead of just generating SFLs. Some work has also been done to explore the information language models capture \cite{ettinger-2020-bert, Laverghetta2021c}, but we note that at present, the ability of TLMs to model semantic fluency has not been explored.

\section{Experiments}

To understand the extent to which TLMs can improve the modeling of SFLs, we set out to establish baselines based on semantic networks, using word association data similar to the approaches cited earlier, and comparing their performance to TLMs'. We use the SNAFU Sample dataset, cleaning it to correct suspected data entry errors (like autocorrecting typos).

\subsection{Experimental Setup}
    Assume participant $u$ generates an SFL $L = \{w_1, ..., w_{|L|}\}$ in response to category cue $C$ (animals, fruits, etc.). Given a function $f$ based on an approach (described below) which takes a context $\mathbf{D_n}$ (a list $[C, w_1, ..., w_{n-1}]$ such that $n \leq |L|$), applies some pre-processing to it, and uses the underlying approach, can $f$ predict $w_n$? We use two methods to describe and score $f$:
    \begin{enumerate}
        \item $\textit{Coverage} = |L \cap W_f| / |L|$ where $W_f$ is the set of words considered by $f$ while making its predictions. We also define \textit{scaled log-likelihood within coverage} as the log-likelihood of each in-coverage item in $L$ according to $f$. In other words, the scaled log-likelihood reflects how likely the list $L$ is to be generated by the function $f$. Since this is defined only in coverage, it depends largely on coverage, and a better scaled log-likelihood does not necessarily mean that a function is better.
        \item \textit{Top-$k$ accuracy} is the percentage of times $w_n$ is present in $f$'s top-\textit{k} predictions. Top-k accuracy is independent of coverage and thus, we can compare different functions based just on their top-k scores.
    \end{enumerate}
    For both metrics, function $f_1$ is said to model human performance better than $f_2$ if it has a higher score. We create multiple functions based on each of the following approaches differing in hyperparameter values. We lemmatize the predictions and only keep nouns (due to the category words given) for these approaches. For simplicity, we also assume that a word will never occur twice in the same SFL. 
    
    \subsubsection{Baseline Approaches (Non-TLM based)}
        We use five non-TLM based approaches as baselines:
        \begin{enumerate}
            \item \textit{Random Baseline}: We use a dataset of $1/3$M most frequent unigrams (single words) on the internet\footnote{\url{https://www.kaggle.com/datasets/rtatman/english-word-frequency}} to find the frequency with which unigrams and bigrams occur. The most likely predictions are chosen from this weighted distribution of unigrams and bigrams with the top-k predictions being the top-k most common words. 
            
            \item \textit{Random Walk on USF Norms}: We approximate the censored random walk algorithm \cite{abbott2015random} on the USF Free Association Norms \cite{Nelson2004}. $p(w_n|C)$ is the number of times $C$ was the cue word and $w_n$ was the response divided by the total number of times $C$ was the cue word. Coverage is determined by words that were responses to $C$ in the USF Norms.
            
            \item \textit{Random Walk on SWOW}: Small World of Words (SWOW) \cite{dedeyne_navarro_perfors_brysbaert_storms_2019} is a word association dataset similar to the USF Norms, so the process is the same as that for USF Norms.
            
            \item \textit{Word2Vec}: We use the Google News 300-dimensional word embeddings model \cite{Mikolov2013a}
            to get the cosine similarity score of a word $w_n$ with all the words in $\mathbf{D_n}$. The average of all these scores becomes $\textit{sc}(w_n|\mathbf{D_n})$. Functions based on Word2Vec take a hyperparameter \textit{context size} ($ct$) which is the maximum number of words preceding $w_n$ in the SFL to consider. These context words are included in order to encourage clustering and switching behaviors (\S \ref{sec:related}), in the hope that higher probability will be assigned to words related to those in close proximity.
            We test four different values for $\textit{ct}$ (0, 1, 3, 5) because our experiments did not show a noticeable difference between $\textit{ct}=5$ and $\textit{ct}>5$. To save runtime, $ct=2,4$ were eliminated too, giving us four Word2Vec based functions.
            
            \item GloVe: Just like Word2Vec, we use a GloVe based word embeddings model glove.6B.300d pre-trained on Wikipedia 2014 + Gigaword 5 \cite{Parker_Graff_Kong_Chen_Maeda_2011}. We use the same \textit{context size} values for GloVe based functions as used for Word2Vec based functions.
            
        \end{enumerate}
    
    \subsubsection{TLM-based approaches}
        We discussed several reasons behind the intuition to use TLMs to model human SFLs in \S \ref{sec:intro}. We perform the MLM task (\S \ref{sec:intro}) on pre-trained TLMs using empirically generated prompts for a category $C$ and a context size $ct$:
        \begin{enumerate}
            \item \textit{The $w_{n-1-ct}$, ..., the $w_{n-1}$, and the \texttt{[MASK]} are examples of Cs.}
            \item \textit{Examples of Cs are the $w_{n-1-ct}$, ..., the $w_{n-1}$, and the \texttt{[MASK]}.}
            \item \textit{The $w_{n-1-ct}$, ..., the $w_{n-1}$, and the \texttt{[MASK]} are the first Cs that come to my mind.}
            \item \textit{The first Cs that come to my mind are the $w_{n-1-ct}$, ..., the $w_{n-1}$, and the \texttt{[MASK]}.}
        \end{enumerate}
        Most of these prompts have the word `the' preceding all the SFL items because, without it, TLMs tended to predict stopwords much more often in our preliminary experiments. Context sizes $\textit{ct}=0,1,3,5$ are tested, as with Word2Vec and GloVe. Each TLM-based function differs in $\textit{ct}$ - prompt pair, giving us 56 functions for each TLM.
        
        Each of these TLMs split the input prompts into tokens such that more than one token is required to encode some words (for example, `blueberry' is encoded by RoBERTa using two tokens). We use a greedy strategy to allow our functions to predict such words. Since one mask is insufficient to predict some words, we also use two consecutive masks for the TLM to allow subwords for each of those masks. A prompt with $ct=1$ would look like ``\textit{Examples of fruits are the strawberry and the \texttt{[MASK][MASK]}.}'' We take the top 100 predictions the TLM made for the first mask and pass a new prompt with each of those predictions replacing the first mask to get 15 predictions for the second mask. Each TLM outputs a softmax distribution over all its tokens corresponding to the mask token. After choosing the top 15 predictions, we scale their probability to add up to 1. These probabilities are multiplied by the previous prediction's probabilities to get a valid probability distribution for the two-mask sequence. Since our function does not know the word we are trying to predict, we generate 3000 one-mask, 1500 two-mask, 400 three-mask, and 100 four-mask predictions for each function (these values were chosen to balance search space size and computation time based on preliminary tests; their effect on performance was minimal because we report top-1 and top-5 scores and these values are well over that range). Since the cumulative probabilities of these four sets of predictions add up to 4, we scale them based on how frequently these words occur in the dataset of 1/3M most frequent words on the internet (note that this is an estimate used to weigh the predictions).
        
        The TLMs we use in this paper are DistilBERT-base-uncased \cite{distilbert}, RoBERTa-Large \cite{liu2019roberta}, and RoBERTa-Med-Small-1M-2 (commonly known as the smallest miniBERTa) \cite{warstadt2020learning}. The models differ in architecture, size, and perhaps most importantly, pre-training data amounts. The smallest miniBERTa is pre-trained on just 1M words, DistilBERT-base-uncased is a smaller version of BERT \cite{Vaswani2017} pre-trained on about 3.4B words, and RoBERTa-Large is pre-trained on approximately 34B words. Since pre-training is the only training these models get before using them in our functions, we can hypothesize that lesser pre-training data (miniBERTa) might lead to poorer performance.

\subsection{Experiment 1: Which approach is the best at modeling SFLs?}
\label{sec:exp1}

    \begin{wrapfigure}{r}{0.5\textwidth}
    \begin{equation}
        \displaystyle
        \mathit{Avg} = \frac{1}{|\mathbf{L}||\mathbf{F}|} \underset{L \in \mathbf{L}}{\sum} \underset{f \in \mathbf{F}}{\sum} S(f,L)
        \label{eq:avg}
    \end{equation}
    \begin{equation}
        \displaystyle
        \mathit{BO} = \frac{1}{|\mathbf{L}|} \underset{f}{\max} \underset{L \in \mathbf{L}}{\sum} S(f,L)
        \label{eq:bo}
    \end{equation}
    \begin{equation}
        \displaystyle
        \mathit{BI} = \frac{1}{|\mathbf{L}|} \underset{L \in \mathbf{L}}{\sum} \underset{f}{\max}\; S(f,L)
        \label{eq:bi}
    \end{equation}
    \end{wrapfigure}
    
    How well do our TLM-based approaches model human SFLs, compared to non-TLM-based approaches? And do the different hyperparameters make the approaches better? Furthermore, if certain approaches and hyperparameter values effectively model human SFLs, do they tell us anything about human cognitive traits? Our experimental setup is designed to be a comparative study: we record the performance (coverage, scaled log-likelihoods, and top-$k$ accuracies) of each of our functions (from all approaches) for each SFL in SNAFU Sample (\S \ref{sec:related}). Let $\mathbf{L}$ be the set of SFLs in SNAFU Sample, and let $S(f,L)$ denote the score (any metric) of function $f$ on SFL $L$. Let $\mathbf{F}$ be the set of all functions for an approach. Since we tested a wide range of hyperparameter value settings (functions) for each approach, we define and report the \textit{approach average (Avg)}, \textit{best overall ($\textit{BO}$)}, and \textit{best individual ($\textit{BI}$)} scores as defined by Equations \ref{eq:avg}, \ref{eq:bo}, and \ref{eq:bi}.

    \begin{table}
    \centering
    \footnotesize
    \begin{tabular}{|c||c|c|c||c|c|c||c|c|c||c|c|c|}
        \hline
        Approach & \multicolumn{3}{c||}{Coverage (\%)} & \multicolumn{3}{c||}{Scaled LL} & \multicolumn{3}{c||}{top-1 accuracy (\%)} & \multicolumn{3}{c|}{top-5 accuracy (\%)} \\
        \cline{2-13}
        {} & \textit{Avg} & \textit{BO} & \textit{BI} & \textit{Avg} & \textit{BO} & \textit{BI} & \textit{Avg} & \textit{BO} & \textit{BI} & \textit{Avg} & \textit{BO} & \textit{BI} \\
        \hline
        \textit{Random} & \textbf{96.4} & \textbf{96.4} & \textbf{96.4} & $-$12.9 & $-$12.9 & $-$12.9 & 0.0 & 0.0 & 0.0 & 0.0 & 0.0 & 0.0 \\
        \textit{USF} & 46.2 & 46.2 & 46.2 & $-$6.2 & $-$6.2 & $-$6.2 & 2.9 & 2.9 & 2.9 & 11.6 & 11.6 & 11.6 \\
        \textit{SWOW} & 71.2 & 71.2 & 71.2 & $-$7.7 & $-$7.7 & $-$7.7 & 4.6 & 4.6 & 4.6 & 16.2 & 16.2 & 16.2 \\
        \textit{Word2Vec} & 59.6 & 65.1 & 67.5 & $-$5.5 & $-$5.5 & $-$5.5 & 2.2 & 3.0 & 4.3 & 8.4 & 10.4 & 12.8 \\
        \textit{GloVe} & 53.6 & 62.8 & 64.1 & $-$\textbf{5.2} & $-$\textbf{4.8} & $-$\textbf{4.7} & 2.0 & 2.9 & 3.8 & 7.8 & 10.8 & 12.6 \\
        \textit{miniBERTa} & 19.7 & 23.9 & 29.4 & $-$9.3 & $-$8.8 & $-$6.8 & 0.0 & 0.1 & 0.2 & 0.1 & 0.3 & 0.7 \\
        \textit{DistilBERT} & 57.8 & 58.9 & 60.5 & $-$6.6 & $-$6.2 & $-$5.3 & 3.5 & 4.9 & 8.6 & 11.6 & 14.7 & 19.8 \\
        \textit{RoBERTa} & 76.2 & 80.1 & 82.6 & $-$6.0 & $-$5.5 & $-$4.9 & \textbf{4.9} & \textbf{6.9} & \textbf{10.3} & \textbf{16.1} & \textbf{20.3} & \textbf{25.1} \\
        \hline
        \hline
        ATC & - & - & - & - & - & - & 7.0 & 9.5 & 20.8 & 19.1 & 23.6 & 47.5 \\
        CA & - & - & - & - & - & - & \textbf{\underline{9.1}} & \textbf{\underline{11.0}} & \textbf{\underline{21.4}} & \textbf{\underline{24.4}} & \textbf{\underline{27.0}} & \textbf{\underline{49.0}} \\
        \hline
    \end{tabular}
    \caption{Performance comparisons of all approaches. The scores for the best static approaches are in bold and the overall best scores are underlined and in bold. Note that larger scaled log-likelihoods and larger accuracies are better.}
    \label{tab:aggregate}
    \end{table}
    
    \begin{table}
    \centering
    \footnotesize
    \begin{tabular}{|c||c|c|c|c|c|c||c|c|c|c|c|c|}
        \hline
        Approach & \multicolumn{6}{c||}{Coverage (\%)} & \multicolumn{6}{c|}{top-5 accuracy (\%)} \\
        \cline{2-13}
        {} & \textit{Fruits} & \textit{Veg} & \textit{Ani} & \textit{Sup} & \textit{Tools} & \textit{Foods} & \textit{Fruits} & \textit{Veg} & \textit{Ani} & \textit{Sup} & \textit{Tools} & \textit{Foods} \\
        \hline
        \textit{USF} & 63.9 & 58.3 & 56.2 & 1.8 & 41.1 & 43.5 & 35.5 & 22.8 & 4.1 & 0.0 & 30.7 & 0.0 \\
        \textit{SWOW} & \textbf{90.6} & 81.5 & 82.2 & 13.0 & 64.2 & 76.0 & \textbf{48.2} & 22.9 & 12.3 & 0.0 & \textbf{33.3} & 0.0 \\
        \textit{Word2Vec} & 90.0 & 83.9 & 74.0 & 0.0 & 59.1 & 72.5 & 15.3 & 17.6 & 13.1 & 0.0 & 10.8 & 7.9 \\
        \textit{GloVe} & 86.9 & 79.0 & 77.5 & 0.0 & 44.8 & 70.7 & 16.9 & 19.3 & 14.3 & 0.0 & 7.5 & 8.1 \\
        \textit{miniBERTa} & 25.6 & 18.2 & 27.3 & 31.4 & 16.2 & 29.0 & 1.3 & 0.1 & 0.1 & 0.6 & 0.1 & 0.6 \\
        \textit{DistilBERT} & 67.7 & 45.5 & 69.8 & 50.6 & 45.3 & 59.7 & 29.5 & 24.1 & 20.5 & 7.3 & 12.0 & 12.9 \\
        \textit{RoBERTa} & 87.9 & \textbf{86.4} & \textbf{82.5} & \textbf{68.3} & \textbf{75.4} & \textbf{81.9} & 30.6 & \textbf{26.7} & \textbf{21.7} & \textbf{12.7} & 22.8 & \textbf{17.7} \\
        \hline
    \end{tabular}
    \caption{Best Overall (BO) coverage and top-5 accuracy (\%) on SFL categories. The scores for the best performing approach on each category are in bold.}
    \label{tab:category}
    \end{table}

    \subsubsection{Results}
    \label{sec:e1-results}
        The top part of Table \ref{tab:aggregate} shows the performance comparison of all approaches. \textit{Random} baseline has a high coverage because the search space is 1/3M most common words on the internet. \textit{RoBERTa-Large}\footnote{To avoid confusion, TLM names are in regular font and approach names are in italics} proves to be the best performing approach when generalized across all users, closely followed by \textit{DistilBERT}. \textit{miniBERTa} is outstandingly poor, performing worse than all baselines, proving our hypothesis about more pre-training data leading to better performance.
        
        On an aggregate level, TLM-based approaches outperform non-TLM-based approaches. Best Overall (\textit{BO}) reports the average scores of the function (approach - hyperparameters combination) that performed the best across all lists. Best Individual (\textit{BI}) picks the best function for each SFL and reports the average scores of this group of functions. \textit{BO} and \textit{BI} show the performance of an approach if we were able to choose the best hyperparameter setting for those approaches.
        \textit{Avg}, \textit{BO}, and \textit{BI} are the same for approaches which do not have different hyperparameter values (multiple functions). Note that best individual (\textit{BI}) is just a theoretical upper limit, as it assumes that we choose the function which is the best for a particular SFL without any means of knowing which one that might be. Thus, practical comparisons must be made with \textit{BO} instead of \textit{BI}.
        
        In order to get a deeper understanding of the strengths (and weaknesses) of TLM-based approaches, we look at how they perform on each individual category (Table \ref{tab:category}). We report coverage and top-$5$ accuracy, which is a common metric for cases where the number of classes is either large or not strictly defined (each SFL category can have many items belonging to it)
        \cite{Luo_2017_ICCV,ravuri2019classification} and more lenient than top-$1$ accuracy. We can see from Table \ref{tab:category} that \textit{SWOW} has a much better performance on `fruits' and `tools', possibly because the smaller and higher quality search space SWOW has as a semantic network is suited for categories that have fewer items belonging to them in general (`fruits' and `tools' have median list lengths of 18 and 16 respectively). However, smaller search space is a double-edged sword and hurts performance on uncommon and wider categories like `supermarket items' and `foods' (median list lengths 35 and 36.5 respectively). The most significant performance gain by TLM-based approaches is on these uncommon categories. Possible reasons for this are larger vocabulary (coverage), better context-awareness (all these TLMs have attention heads), and higher processing capabilities. Note that we never fine-tuned these TLMs, and we are using just the pre-trained versions, so any performance reflected here is due to learning during pre-training.
        

\subsection{Experiment 2: Can TLMs identify individual differences?}
    
    
    
    Table \ref{tab:aggregate} clearly shows that large TLMs outperform the other language model types we considered (except miniBERTa, which is significantly smaller than the other TLMs listed). However, such aggregate evaluations can sometimes hide interesting information. For example, is it the case that certain functions better model \textit{individual} SFLs? If so, this might suggest further ways of generating exploratory hypotheses, in line with the idea of hyperparameter hypothesization (\S \ref{sec:CMs})---e.g., if one prompt type seems to model lists generated by one person best, whereas a different prompt type best models lists generated by another, then this may suggest that there is a qualitative difference in how those two individuals ``query'' their memories when performing this task which is captured by the different prompt styles.
    
    Experiment 2, therefore, aims to take first steps in exploring the plausibility of these ideas. We start by examining the extent to which specific functions (each of which, recall, consist of a model + hyperparameter values) can adapt to individuals. Our task is conceptualized as follows: Given an individual who is performing the SFT, and a machine that is observing the list being generated one item at a time, how quickly can the machine adapt to that individual and predict what items the individual will list next?
    
    We refer to the functions compared in Experiment 1 as \textit{static}, as they each have fixed hyperparameter values. They are compared to two \textit{adaptive} functions:
    \begin{itemize}
        \item Adapt-Then-Change (ATC): Chooses the static function performing the best on the first $x$ items of the SFL, and applies that chosen function to the rest of the SFL. Performance is averaged across all the items in the latter part of the list.
            
        \item Continuous Adaptation (CA): Uses a sliding window of size $x$ and chooses the best performing function for all the items in this window. It then applies that chosen function to the next item. After this, the window slides one item to the right. Whereas ATC only makes one decision about which function to use per list, CA makes that decision for each list item (starting from the $x^{th}$ item).
        
    \end{itemize}
    We do not calculate coverage and scaled log-likelihoods in coverage for adaptive approaches because they are not new approaches, just strategies to choose one hyperparameter setting for one of the existing approaches.
    
    \subsubsection{Results}
        \begin{wrapfigure}{r}{0.6\textwidth}
        \centering
            \includegraphics[width=0.55\textwidth]{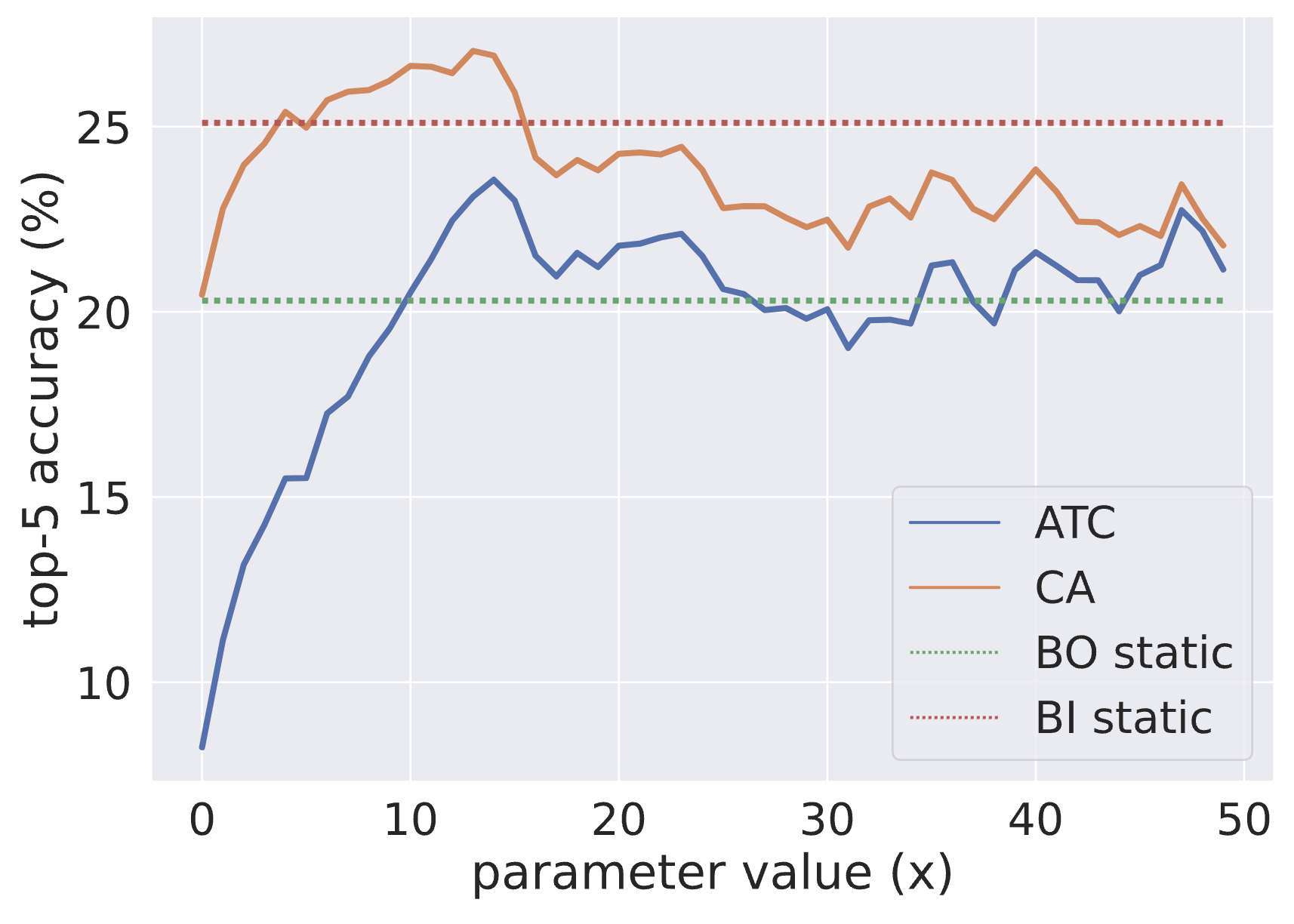}
            \caption{Top-$5$ accuracy comparison of adaptive and static approaches.}
            \label{fig:top5}
        \end{wrapfigure}
        
        From Table \ref{tab:aggregate}, adaptive approaches clearly outperform the static approaches, indicating that a smart way to choose hyperparameters can drastically improve prediction accuracy. Essentially, this means that based on just $w_1, ..., w_{n-1}$ and without knowing $w_n$, we are able to find a set of hyperparameters that can be used to predict the next word with about $27\%$ accuracy. Considering that this is across all categories and across all individuals, and that the search space for this task is not strictly bounded, these results exceed expectations. The adaptive approaches do not have hyperparameters of their own so \textit{BO} and \textit{BI} in Table \ref{tab:aggregate} are calculated based on different values of $x$. 
        
        \begin{table}
        \centering
        \footnotesize
            \begin{tabular}{|c||c|c|c|c|c|c|c|}
                \hline
                {} & \textit{USF} & \textit{SWOW} & \textit{Word2Vec} & \textit{GloVe} & \textit{miniBERTa} & \textit{DistilBERT} & \textit{RoBERTa} \\
                \hline
                ATC & 9.5 & 10.4 & 7.9 & 2.3 & 0.1 & 29.0 & \textbf{40.8} \\
                \hline
                CA & 12.9 & 15.9 & 8.9 & 2.7 & 0.4 & 28.2 & \textbf{31.0} \\
                \hline
            \end{tabular}
            \caption{Percentage of times each static approach was chosen by an adaptive approach.}
            \label{tab:choose}
        \end{table}
        
        The comparison of top-$5$ accuracies between the adaptive approaches and the best static approach (\textit{RoBERTa-Large}) is shown in Figure \ref{fig:top5}. ATC outperforms \textit{BO} static for certain values of $x$, and CA outperforms \textit{BO} static almost always. This is expected because CA adapts to the SFL items continuously while ATC adapts just once. As discussed in \S \ref{sec:e1-results}, \textit{BI} is more of a theoretical maximum, due to the fact that it is selected from all functions for each SFL after we already know it does best on that SFL. Nonetheless, CA still impressively outperforms \textit{BI} static for certain values of $x$.
        
\section{Conclusion, Limitations, and Future Work}

    To our knowledge, we are the first to use transformer-based LMs to model semantic fluency lists. We do so in the hopes of advancing the computational modeling of human semantic knowledge and memory retrieval processes. To that end, we used RoBERTa-Large, DistilBERT-base, and miniBERTa-med-small to model semantic fluency lists and found that RoBERTa-Large and DistilBERT-base consistently outperformed other non-transformer based approaches. We hypothesize that miniBERTa-med-small's poor performance is due to its smaller pre-training corpus. DistilBERT-base's performance suggests that the associative semantic information needed to model human SFLs starts getting imparted as the size of the training corpus increases from 1M words to about 3.4B. Future work can attempt to estimate the size of training corpus in this range where TLMs start outperforming baseline approaches and whether larger training corpora would hurt performance. Hopefully, using upcoming TLMs with a wider variety of prompts inspired by this paper, perhaps using a technique closer to prompt fine-tuning \cite{fichtel2021prompt} will give better results.
    
    Furthermore, we took a first step in exploring the ability of TLMs to determine individual differences in retrieval behaviors in the SFT. Our results suggest that an adaptive approach works best, in some cases even outperforming an oracular baseline. However, we do not yet know if the function and hyperparameter choices that our adaptive approaches make reflect stable individual cognitive traits or behaviors. Answering this question will be the focus of future work, for which the present work has laid an important foundation.
    
    It is possible that fine-tuning on some subset of the SNAFU Sample dataset may yield better likelihoods or top-k scores. Likewise, it may be trivial to fine-tune RoBERTa to output all instances of a given category, or to train a LM to simply enumerate all known hyponyms of a category word. But the goal of this work, as described in \S \ref{sec:CMs}, is primarily to match and then produce hypotheses for cognitive processes. As such, it was not our goal to simply create an exhaustive item list generator; rather, we want to emulate how people generate SFLs.
    
    In using transformer-based LMs to model human response patterns, it should be noted that we are not taking into account how many other psychological and cognitive constructs factor in to the complex retrieval processes involved in SFT. Although the method we describe here is designed to compare individual linguistic retrieval strategies, it is unclear what exactly it tells us about how performance on semantic fluency tasks relates to individuals' executive functioning and self-regulation skills, which fluency tasks are often employed to study
    \cite{whiteside2016,aita2019}. Rather, the work here is the first step in our proposed \textit{hyperparameter hypothesization} strategy (\S \ref{sec:CMs}), which we propose here for the first time and believe contributes to the present symposium's goals.
    

\section{Acknowledgements}
Part of this research was sponsored by the DEVCOM Analysis Center and was accomplished under Cooperative Agreement Number W911NF-22-2-0001. The views and conclusions contained in this document are those of the authors and should not be interpreted as representing the official policies, either expressed or implied, of the Army Research Office or the U.S. Government. The U.S. Government is authorized to reproduce and distribute reprints for Government purposes notwithstanding any copyright notation herein.

\bibliographystyle{unsrtnat}
\bibliography{ijcai22, john}

\end{document}